\documentclass[letterpaper, 10 pt, conference]{ieeeconf}  

\IEEEoverridecommandlockouts                              

\usepackage{pifont}
\usepackage{url}            
\usepackage{booktabs}       
\usepackage{amsfonts}       
\usepackage{nicefrac}       
\usepackage{microtype}      
\usepackage{xcolor}         
\usepackage{listings}
\usepackage{array}
\usepackage{caption}
\usepackage{amsmath}
\usepackage{amssymb}
\usepackage{adjustbox}
\usepackage{tcolorbox}
\usepackage{multirow}
\usepackage{makecell} 
\usepackage{tabularx}
\usepackage{graphicx}
\usepackage{cuted}
\usepackage{multicol}

\definecolor{linkblue}{HTML}{4286F5}
\usepackage{placeins}

\usepackage[
    colorlinks=true,
    linkcolor=linkblue,
    citecolor=linkblue,
    urlcolor=linkblue,
    pdfborder={0 0 0}
]{hyperref}
\usepackage{colortbl}

\title{\LARGE \bf XPolicyLab: A Unified Standard and Open Ecosystem\\for Robot Policy Evaluation and Deployment}

\author{
\parbox{0.9\textwidth}{
\centering
\fontsize{8.5pt}{10pt}\selectfont
\textbf{\hyperref[sec:contributor_list]{XPolicyLab Contributors}}, Leading Institutions: MMLab@HKU \& THU\\[4pt]
Website: \href{https://xpolicylab.github.io/}{https://xpolicylab.github.io/}
}\\[-10pt]
}

\begin{document}

\maketitle
\thispagestyle{empty}
\begin{strip}
\centering
\vspace{-3em}
\includegraphics[width=\linewidth]{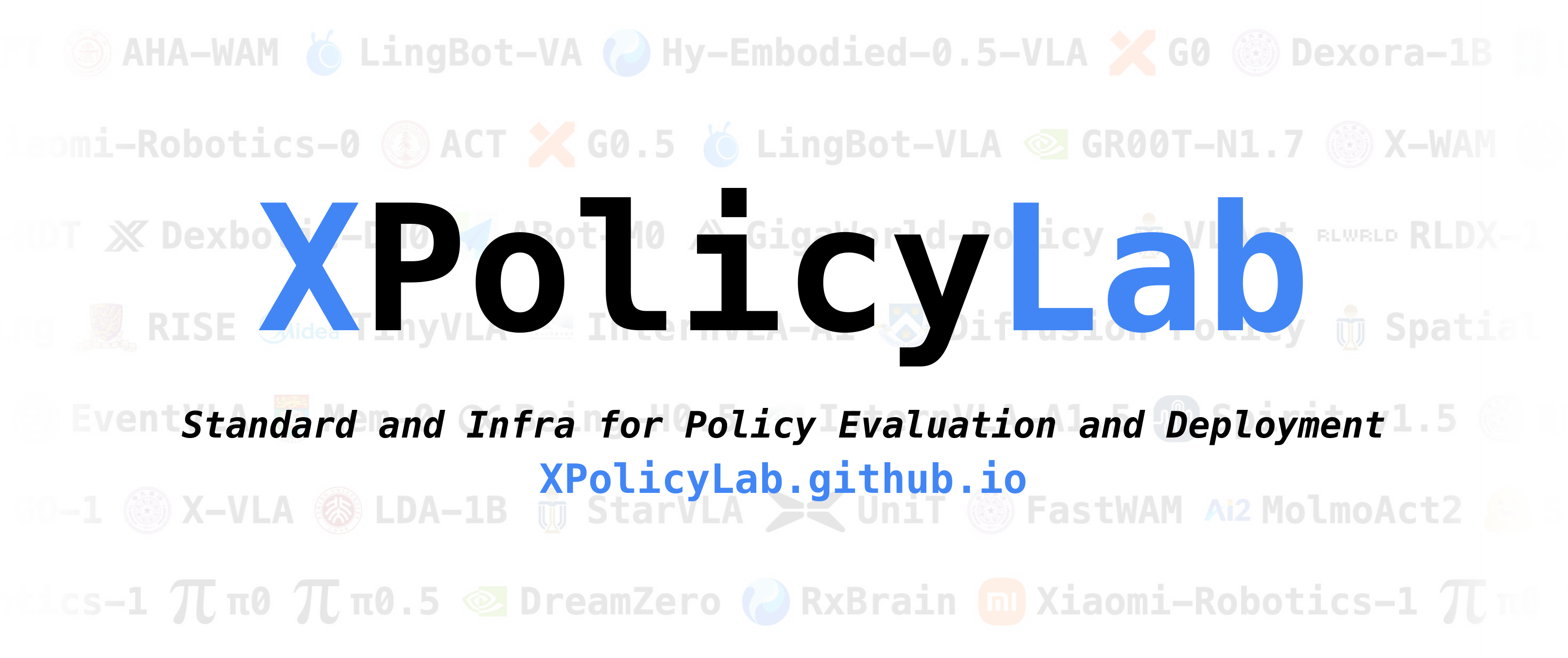}
\captionof{figure}{%
    \textbf{Overview of XPolicyLab.} A standard and infrastructure for robot policy evaluation and deployment.%
}
\label{fig:teaser}
\end{strip}


\begin{abstract}
Robot policy evaluation and deployment remain fragmented by model-specific software dependencies, data representations, and runtime interfaces, so that connecting $N$ policies to $M$ evaluation environments requires $O(NM)$ separate integrations.
We present \textbf{XPolicyLab}, a unified standard and open ecosystem that reduces this cost to $O(N{+}M)$.
XPolicyLab specifies common observation, action, and trajectory schemas together with a minimal adapter interface for observation updates, action prediction, batched execution, and episode reset, while a dependency-isolated client/server architecture separates policy inference from environment execution, so that each side retains its native software stack and may run locally or remotely.
The ecosystem integrates \textbf{42 robot policies} and standardizes their installation, debugging, serving, and evaluation workflows.
Across these adapters, model-specific code varies by an order of magnitude while the environment-facing loop stays within a few lines of a fixed reference, confirming that the contract confines heterogeneity to the policy side.
In a controlled study, conforming to the standard reduces the integration effort of a representative policy from over five hours to two hours, and packaged agent skills reduce it further to thirty minutes.
The same adapters serve RoboTwin, RoboDojo simulation, and standardized real-robot evaluation through one interface.
XPolicyLab is released as shared infrastructure for reproducible policy comparison and standardized deployment across simulation and physical platforms. Project website: \href{https://xpolicylab.github.io/}{https://xpolicylab.github.io/}.
\end{abstract}

\section{Introduction}

Generalist robot policies, including vision-language-action (VLA) models and other embodied foundation models, have rapidly expanded in capability and task coverage~\cite{kim2024openvla,intelligence2025pi_,intelligence2026pi,team2024octo,ye2026world,cai2026xiaomi,zhang2026hy,chen2025g3flow}.
Their evaluation and deployment, however, remain constrained by fragmented system stacks~\cite{yang2026towards}.
Different policies assume incompatible software dependencies, observation and action representations, checkpoint conventions, and runtime interfaces, so deploying a policy on a new benchmark, simulator, or physical robot requires substantial platform-specific engineering.
Observation preprocessing, action conversion, control loops, and runtime orchestration are repeatedly reimplemented, which increases integration effort and, more damagingly, silently re-fixes conventions such as camera naming, channel order, and gripper scaling, so that reported results become difficult to compare across policies.
Stated as a systems problem, connecting $N$ policies to $M$ evaluation environments currently costs $O(NM)$ separate integrations.

Substantial progress has been made in robot data collection, dataset standardization, and training infrastructure~\cite{cadene2026lerobot,liu2024fastumi,khazatsky2024droid,community2026starvla,yu2026rlinf}, which lowers the cost of producing a checkpoint.
The complementary boundary, between a trained policy and the environments that must execute it, remains largely unaddressed.
A reusable solution there must absorb heterogeneous model architectures and dependencies while exposing a stable interface to simulators, benchmark clients, and physical robots.
In particular, it must support stateful policies, action chunking, batched evaluation, local and remote execution, and reliable communication, without coupling policy implementations to environment-specific software stacks.

\begin{figure*}[t]
    \centering
    \includegraphics[width=1.0\linewidth]{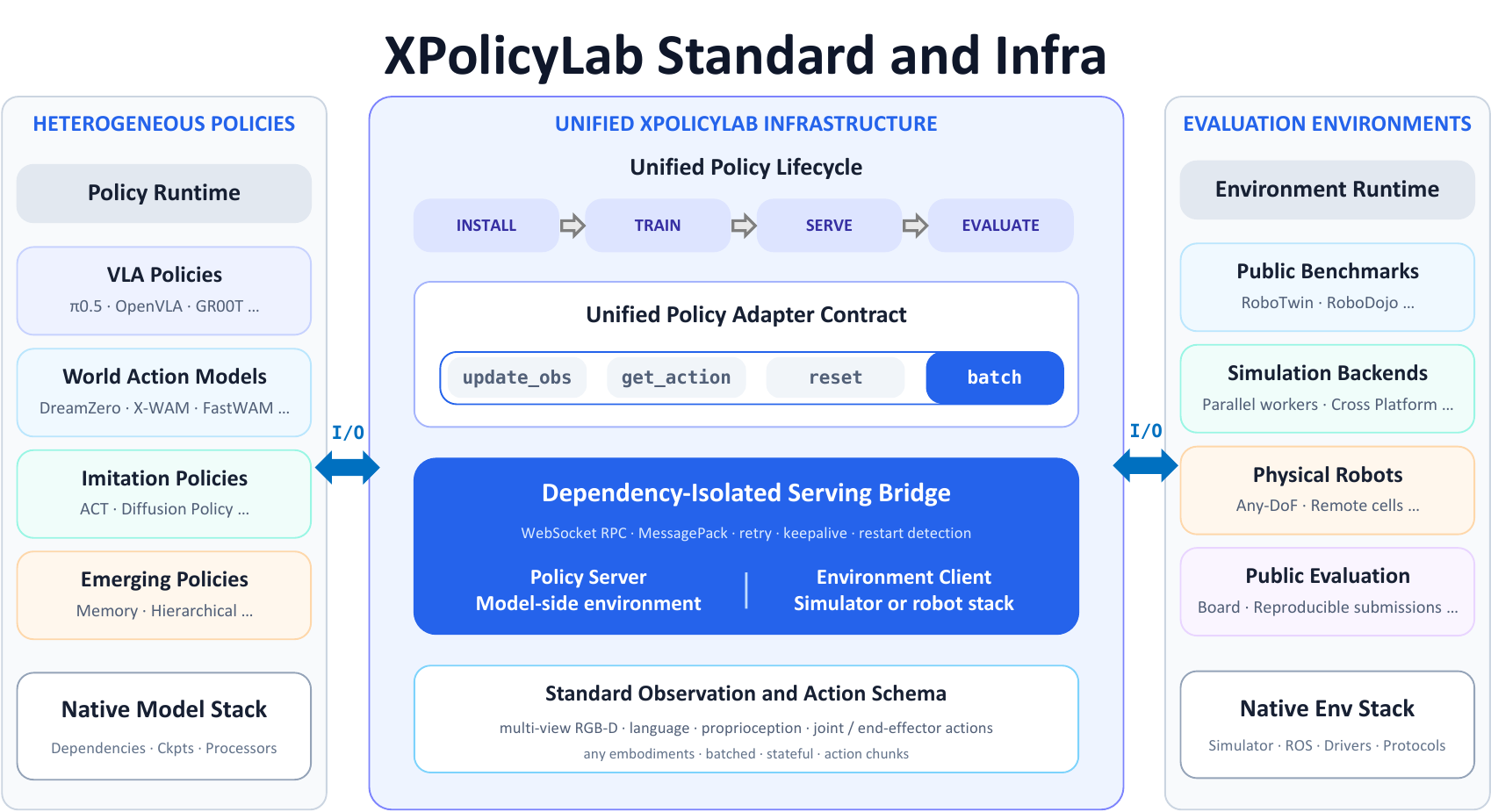}
    \caption{
    \textbf{Infrastructure of XPolicyLab.}
    Heterogeneous policy runtimes (left) keep their native dependencies, checkpoints, and processing pipelines, while evaluation environments (right) keep their simulator and robot stacks.
    XPolicyLab (center) joins the two through a unified lifecycle from installation to evaluation, a minimal adapter contract, standardized observation and action schemas, and a dependency-isolated serving bridge.
    A single adapter therefore serves public benchmarks, simulation backends, and physical robots, under local or remote execution and batched, stateful inference.
    }
    \label{fig:xpolicylab_overview}
    \vspace{-10pt}
\end{figure*}

We present \textbf{XPolicyLab},\footnote{\url{https://github.com/XPolicyLab/XPolicyLab}} a unified standard and open ecosystem for robot policy evaluation and deployment, illustrated in Fig.~\ref{fig:xpolicylab_overview}.
XPolicyLab specifies common observation, action, and trajectory representations, together with a lightweight adapter contract for policy initialization, observation updates, action prediction, batched execution, and episode reset, so that each integrated policy retains its native implementation, dependencies, checkpoints, and training procedure.
A dependency-isolated serving architecture separates policy inference from environment execution through a common communication layer, allowing policy servers and environment clients to run in independent software environments, either on the same machine or across a network.
Around this core, XPolicyLab standardizes the workflows for installation, data conversion, training, serving, debugging, and evaluation, and packages the conformance procedure as machine-readable agent skills that a coding agent executes to scaffold, implement, and validate a new adapter.

As of August 2026, XPolicyLab integrates \textbf{42 robot policies}, spanning VLA models, world-action models, diffusion-based visuomotor policies, memory-augmented policies, and conventional imitation-learning baselines.
By separating model-facing adapters from environment-facing clients, a single adapter is reused across benchmark ecosystems, simulation backends, and physical robot evaluation systems without modifying the underlying policy implementation.
We demonstrate this portability through deployments on RoboTwin~\cite{chen2025robotwin} and RoboDojo~\cite{chen2026robodojo}, including standardized physical evaluation through RoboDojo-RealEval, for which XPolicyLab also provides the policy submission and execution workflow of the official public leaderboards.

Our main contributions are summarized as follows:
\begin{enumerate}
    \item We introduce a unified contract between robot policies and evaluation environments that standardizes observation and action representations, action prediction, batched execution, and episode management while preserving policy-specific implementations.

    \item We develop a dependency-isolated serving architecture that decouples policy inference from environment execution and supports local and remote deployment across simulators, benchmark clients, and physical robot systems.

    \item We package the conformance procedure as machine-readable agent skills, so that scaffolding, implementing, and auditing a new adapter against the standard becomes a guided and checkable procedure rather than a documentation-reading exercise.

    \item We establish an open ecosystem comprising \textbf{42 robot policies} across diverse policy families, and evaluate it in terms of policy coverage, the locality of integration complexity, integration effort, and adapter reuse, with deployments on RoboTwin~\cite{chen2025robotwin} and RoboDojo~\cite{chen2026robodojo}.
\end{enumerate}

\section{Related Work}

\subsection{Robot Data and Training Infrastructure}

Large-scale datasets and open-source frameworks have improved the availability of robot demonstrations and the reproducibility of policy training.
DROID and BridgeData provide diverse real-world manipulation data~\cite{khazatsky2024droid,ebert2021bridge}, FastUMI supports embodiment-agnostic data collection~\cite{liu2024fastumi}, LeRobot standardizes dataset processing and policy training~\cite{cadene2026lerobot}, and RLinf provides scalable infrastructure for policy optimization and execution~\cite{yu2026rlinf}.
These systems target data acquisition, dataset organization, and model optimization, and therefore lower the cost of producing a checkpoint.
XPolicyLab addresses the complementary boundary of how an already-trained policy is integrated, served, evaluated, and deployed across heterogeneous environment stacks.

\subsection{Robot Policy Evaluation Platforms}

Simulation benchmarks such as RLBench, ManiSkill2, LIBERO, RMBench, and UniVTAC provide controlled environments for evaluating robotic skills and generalization~\cite{james2019rlbench,maniskill2,liu2023libero,chen2026rmbench,chen2026univtac}.
RoboTwin extends this setting to diverse bimanual manipulation tasks~\cite{mu2025robotwin,chen2025robotwin,chen2025benchmarking}, while RoboDojo combines capability-oriented simulation with standardized physical evaluation~\cite{chen2026robodojo}.
Beyond simulation, RoboArena and RoboChallenge coordinate distributed real-robot evaluation across institutions and hardware~\cite{atreya2025roboarena,yakefu2025robochallenge}.
These platforms define task semantics, environment interfaces, evaluation protocols, and performance metrics; that is, they specify what is evaluated and under which conditions.
XPolicyLab does not replace them, and instead supplies the policy-side abstraction that lets heterogeneous policies reach any of them through one interface, so that no benchmark needs to maintain its own model-specific serving stack.

\subsection{Robot Policy Deployment}

Robot policies such as OpenVLA, $\pi_0$, and GR00T ship inference and deployment code tailored to their respective architectures, dependencies, observation formats, and action representations~\cite{kim2024openvla,black2024pi_0,bjorck2025gr00t}.
Each such stack is faithful to one model, which is precisely why deploying a policy across multiple benchmarks, simulators, and physical robots requires repeated adaptation of preprocessing pipelines, action conversion, runtime environments, and control loops, and why the community pays the $O(NM)$ cost identified above.
XPolicyLab removes this pairwise dependency through standardized observation and action schemas, a minimal adapter interface, and dependency-isolated serving.
Unlike a unified modeling framework, it prescribes no network architecture, training objective, action decoder, or temporal horizon, which is what allows adapters to be reused across evaluation environments while model-specific implementations remain intact.

\section{XPolicyLab}
\label{sec:method}

XPolicyLab adopts a policy-centric abstraction that decouples policy computation from environment execution.
Its central principle is that each policy exposes a stable external contract while retaining control over its architecture, dependencies, checkpoints, and inference procedure, so that heterogeneous benchmarks, simulators, and physical robots can invoke any policy through a common execution pathway.

\subsection{System Overview}
\label{sec:overview}

XPolicyLab comprises four functional components: an environment backend that executes tasks, an environment client that manages the interaction loop, a communication layer that transfers observations and actions, and a policy server that hosts the model and its adapter.
Cross-cutting workflow tooling standardizes installation, configuration, debugging, training, serving, and evaluation.
The policy server executes in an independent software environment and may reside either on the same machine as the environment client or on a remote compute node.

The architecture establishes two stable boundaries.
A \emph{semantic boundary} specifies the shared observation and action schemas, whereas an \emph{execution boundary} defines communication between policy servers and environment clients.
For an observation $\mathbf{o}_t$, an optional internal state $\mathbf{s}_t$, and an action horizon $H$, policy inference can be expressed as
\begin{equation}
\mathbf{a}_{t:t+H-1}
=
g_{\mathrm{out}}
\left(
\pi_{\theta}
\left(
g_{\mathrm{in}}(\mathbf{o}_t),\mathbf{s}_t
\right)
\right),
\label{eq:policy_interface}
\end{equation}
where $g_{\mathrm{in}}$ and $g_{\mathrm{out}}$ map between the shared schema and the policy-native representations.
XPolicyLab standardizes these external mappings without constraining the internal forms of $\pi_{\theta}$, $g_{\mathrm{in}}$, or $g_{\mathrm{out}}$.

\subsection{Policy Adapter Contract}
\label{sec:adapter}

A policy adapter implements four core operations.
Model construction loads the checkpoint, processors, and runtime configuration.
\texttt{update\_obs} receives a standardized observation and updates the policy context, while \texttt{get\_action} returns an action chunk under the shared schema.
\texttt{reset} clears episode-specific state, which is necessary for recurrent policies, temporal aggregation methods, and models with explicit memory.
The corresponding \texttt{update\_obs\_batch} and \texttt{get\_action\_batch} operations support parallel evaluation and align predictions with the currently active environment instances.

The contract deliberately leaves the network architecture, action decoder, prediction horizon, and training framework unspecified, so that each adapter only translates between the shared interface and the policy's native inference API.
The same evaluation loop therefore accommodates autoregressive VLA models, diffusion-based visuomotor policies, world-action models, and conventional imitation-learning methods.
Integration follows a common lifecycle from installation and data preparation to configuration, serving, and evaluation, with data conversion and training omitted for declared evaluation-only integrations that ship pretrained checkpoints.

\subsection{Dependency-Isolated Serving}
\label{sec:serving}

Robot policies frequently depend on software stacks that conflict with simulator or robot-control environments.
XPolicyLab addresses this through process-level isolation: the policy server loads the model in its native runtime, while the environment client remains in the software environment required by the benchmark, simulator, or robot driver.
The two processes communicate over a WebSocket protocol with MessagePack serialization extended for arrays, using the message set \texttt{HELLO}, \texttt{PREPARE\_CASE}, \texttt{RESET}, \texttt{INFER}, \texttt{CALL}, \texttt{TRIAL\_END}, \texttt{HEARTBEAT}, and \texttt{CLOSE}, each paired with an acknowledgement or result frame.

Evaluation follows the stateful loop
$\texttt{reset}\rightarrow
(\texttt{observe}\rightarrow\texttt{update\_obs}\rightarrow
\texttt{get\_action}\rightarrow\texttt{execute})^{*}$.
For action-chunking policies, the environment client determines how many predicted actions to execute before requesting a new chunk, which keeps that decision on the side that knows the control rate.
The same protocol carries batched inference over multiple environment instances and distributed execution, allowing inference to run on a dedicated GPU server while simulation or physical control remains close to the environment.

Reliability is part of the contract, because a naive retry corrupts policy state.
Requests carry unique identifiers and completed responses are cached, so a retry issued after a transient reconnection returns the cached result instead of running inference a second time; retries of a logical request therefore reuse its identifier.
Each server reports an instance identifier during the handshake, and a change of that identifier is treated as fatal to the current trial, because a restarted server has lost the episode state on which the trajectory depends.
Configurable connection budgets, keepalive checks, and inference timeouts accommodate long checkpoint loading times and variable inference latency.

\newcommand{\XPLPolicyPNG}[4]{%
  \href{#1}{%
    \raisebox{-0.20\height}{%
      \includegraphics[
        height=2.5ex,
        keepaspectratio
      ]{figure/robodojo-logos/#2}%
    }%
    \hspace{2pt}#3%
  }~\cite{#4}%
}

\newcommand{\XPLPolicyNoLogo}[3]{%
  \href{#1}{#2}~\cite{#3}%
}

\begin{table*}[t]
\centering
\caption{
Robot policies supported by XPolicyLab as of August 8, 2026.
Links point to the corresponding adapter implementations.
}
\label{tab:policy_ecosystem}

\setlength{\tabcolsep}{4pt}
\renewcommand{\arraystretch}{1.8}

\resizebox{\textwidth}{!}{%
\begin{tabular}{*{6}{c}}
\toprule

\XPLPolicyPNG
{https://github.com/XPolicyLab/XPolicyLab/tree/main/policy/A1}
{A1.png}
{A1}
{zhang2026a1}
&
\XPLPolicyPNG
{https://github.com/XPolicyLab/XPolicyLab/tree/main/policy/AHA_WAM}
{AHA-WAM.png}
{AHA-WAM}
{cai2026aha}
&
\XPLPolicyPNG
{https://github.com/XPolicyLab/XPolicyLab/tree/main/policy/Abot_M0}
{ABot-M0.png}
{ABot-M0}
{yang2026abot}
&
\XPLPolicyPNG
{https://github.com/XPolicyLab/XPolicyLab/tree/main/policy/ACT}
{ACT.png}
{ACT}
{zhao2023learning}
&
\XPLPolicyPNG
{https://github.com/XPolicyLab/XPolicyLab/tree/main/policy/Being_H05}
{being-beyond.png}
{Being-H0.5}
{luo2026being}
&
\XPLPolicyPNG
{https://github.com/XPolicyLab/XPolicyLab/tree/main/policy/Dexbotic_DM0}
{DM0.png}
{Dexbotic-DM0}
{yu2026dm0}
\\

\XPLPolicyPNG
{https://github.com/XPolicyLab/XPolicyLab/tree/main/policy/Dexora_1B}
{Dexora-1B.png}
{Dexora-1B}
{zhang2026dexora}
&
\XPLPolicyPNG
{https://github.com/XPolicyLab/XPolicyLab/tree/main/policy/DP}
{columbia.png}
{Diffusion Policy}
{chi2025diffusion}
&
\XPLPolicyPNG
{https://github.com/XPolicyLab/XPolicyLab/tree/main/policy/DreamZero}
{GROOT-N1.7.png}
{DreamZero}
{ye2026world}
&
\XPLPolicyPNG
{https://github.com/XPolicyLab/XPolicyLab/tree/main/policy/EventVLA}
{EventVLA.png}
{EventVLA}
{yang2026eventvla}
&
\XPLPolicyPNG
{https://github.com/XPolicyLab/XPolicyLab/tree/main/policy/FastWAM}
{Fast-WAM.png}
{FastWAM}
{yuan2026fast}
&
\XPLPolicyPNG
{https://github.com/XPolicyLab/XPolicyLab/tree/main/policy/G05}
{GalaxeaVLA.png}
{G0.5}
{galaxea2026g05}
\\

\XPLPolicyPNG
{https://github.com/XPolicyLab/XPolicyLab/tree/main/policy/GalaxeaVLA}
{GalaxeaVLA.png}
{G0}
{jiang2025galaxea}
&
\XPLPolicyPNG
{https://github.com/XPolicyLab/XPolicyLab/tree/main/policy/GigaWorldPolicy}
{GigaWorld-Policy-0.png}
{GigaWorld-Policy}
{ye2026gigaworld}
&
\XPLPolicyPNG
{https://github.com/XPolicyLab/XPolicyLab/tree/main/policy/GO1}
{GO-1.png}
{GO-1}
{bu2025agibot}
&
\XPLPolicyPNG
{https://github.com/XPolicyLab/XPolicyLab/tree/main/policy/GR00T_N17}
{GROOT-N1.7.png}
{GR00T-N1.7}
{bjorck2025gr00t}
&
\XPLPolicyPNG
{https://github.com/XPolicyLab/XPolicyLab/tree/main/policy/H_RDT}
{H-RDT.png}
{H-RDT}
{bi2026h}
&
\XPLPolicyPNG
{https://github.com/XPolicyLab/XPolicyLab/tree/main/policy/Hy_Embodied_05_VLA}
{Hy-Embodied-0.5-VLA.png}
{Hy-Embodied-0.5-VLA}
{zhang2026hy}
\\

\XPLPolicyPNG
{https://github.com/XPolicyLab/XPolicyLab/tree/main/policy/InternVLA_A1}
{InternVLA-A1.png}
{InternVLA-A1}
{cai2026internvla}
&
\XPLPolicyPNG
{https://github.com/XPolicyLab/XPolicyLab/tree/main/policy/InternVLA_A1_5}
{InternVLA-A1.5.png}
{InternVLA-A1.5}
{ma2026internvla}
&
\XPLPolicyPNG
{https://github.com/XPolicyLab/XPolicyLab/tree/main/policy/LDA_1B}
{LDA-1B.png}
{LDA-1B}
{lyu2026lda}
&
\XPLPolicyPNG
{https://github.com/XPolicyLab/XPolicyLab/tree/main/policy/LingBot_VA}
{LingBot-VLA.png}
{LingBot-VA}
{li2026causalworldmodelingrobot}
&
\XPLPolicyPNG
{https://github.com/XPolicyLab/XPolicyLab/tree/main/policy/LingBot_VLA}
{LingBot-VLA.png}
{LingBot-VLA}
{wu2026pragmatic}
&
\XPLPolicyPNG
{https://github.com/XPolicyLab/XPolicyLab/tree/main/policy/Mem_0}
{hku.png}
{Mem-0}
{chen2026rmbench}
\\

\XPLPolicyPNG
{https://github.com/XPolicyLab/XPolicyLab/tree/main/policy/MolmoACT2}
{MolmoAct2.png}
{MolmoAct2}
{fang2026molmoact2}
&
\XPLPolicyPNG
{https://github.com/XPolicyLab/XPolicyLab/tree/main/policy/OpenVLA_OFT}
{OpenVLA-OFT.png}
{OpenVLA-OFT}
{kim2024openvla}
&
\XPLPolicyPNG
{https://github.com/XPolicyLab/XPolicyLab/tree/main/policy/Pi_0}
{Pi-0.png}
{$\pi_0$}
{black2024pi_0}
&
\XPLPolicyPNG
{https://github.com/XPolicyLab/XPolicyLab/tree/main/policy/Pi_05}
{Pi-05.png}
{$\pi_{0.5}$}
{intelligence2025pi_}
&
\XPLPolicyPNG
{https://github.com/XPolicyLab/XPolicyLab/tree/main/policy/starVLA}
{Spatial_Forcing.png}
{VLAct}
{ye2026starvla}
&
\XPLPolicyPNG
{https://github.com/XPolicyLab/XPolicyLab/tree/main/policy/RDT_1B}
{RDT.png}
{RDT-1B}
{liu2025rdt}
\\
\XPLPolicyPNG
{https://github.com/XPolicyLab/XPolicyLab/tree/main/policy/RISE}
{cuhk.png}
{RISE}
{yang2026rise}
&
\XPLPolicyPNG
{https://github.com/XPolicyLab/XPolicyLab/tree/main/policy/SmolVLA}
{SmolVLA.png}
{SmolVLA}
{shukor2025smolvla}
&
\XPLPolicyPNG
{https://github.com/XPolicyLab/XPolicyLab/tree/main/policy/Spatial_Forcing}
{Spatial_Forcing.png}
{Spatial Forcing}
{li2026spatial}
&
\XPLPolicyPNG
{https://github.com/XPolicyLab/XPolicyLab/tree/main/policy/Spirit_v15}
{Spirit_v1.5.png}
{Spirit v1.5}
{spiritai2026spiritv15}
&
\XPLPolicyPNG
{https://github.com/XPolicyLab/XPolicyLab/tree/main/policy/starVLA}
{StarVLA.png}
{StarVLA}
{community2026starvla}
&
\XPLPolicyPNG
{https://github.com/XPolicyLab/XPolicyLab/tree/main/policy/TinyVLA}
{TinyVLA.png}
{TinyVLA}
{wen2025tinyvla}
\\

\XPLPolicyPNG
{https://github.com/XPolicyLab/XPolicyLab/tree/main/policy/X_VLA}
{X-VLA.png}
{X-VLA}
{zheng2026x}
&
\XPLPolicyPNG
{https://github.com/XPolicyLab/XPolicyLab/tree/main/policy/X_WAM}
{X-WAM.png}
{X-WAM}
{guo2026unified}
&
\XPLPolicyPNG
{https://github.com/XPolicyLab/XPolicyLab/tree/main/policy/Xiaomi_Robotics_0}
{Xiaomi-Robotics-0.png}
{Xiaomi-Robotics-0}
{cai2026xiaomi}
&
\XPLPolicyPNG
{https://github.com/XPolicyLab/XPolicyLab/tree/main/policy/Xiaomi_Robotics_1}
{Xiaomi-Robotics-1.png}
{Xiaomi-Robotics-1}
{team2026xiaomi}
&
\XPLPolicyPNG
{https://github.com/XPolicyLab/XPolicyLab/tree/main/policy/UniT}
{XPeng.png}
{UniT}
{uniT}
&
\XPLPolicyPNG
{https://github.com/XPolicyLab/XPolicyLab/tree/main/policy/RxBrain}
{Hy-Embodied-0.5-VLA.png}
{RxBrain}
{liang2026rxbrain}
\\

\bottomrule
\end{tabular}%
}
\end{table*}

\subsection{Standardized Observation and Action Schemas}
\label{sec:schema}

XPolicyLab represents an observation as
\begin{equation}
\mathbf{o}_t =
\left\{
\mathbf{v}_t,\mathbf{q}_t,\mathbf{p}_t,\ell,\mathbf{m}_t
\right\},
\label{eq:observation_schema}
\end{equation}
where $\mathbf{v}_t$ denotes visual inputs, $\mathbf{q}_t$ robot joint states, $\mathbf{p}_t$ Cartesian poses, $\ell$ the language instruction, and $\mathbf{m}_t$ optional metadata.
Visual inputs are indexed by camera name and may contain RGB images, depth maps, and camera calibration parameters.
Robot states support single-arm and bimanual embodiments, including arm joints, gripper states, and end-effector poses, and Cartesian poses follow the convention $[x,y,z,q_w,q_x,q_y,q_z]$.
All fields are optional except those required by the evaluated policy.

Actions follow the same embodiment-aware organization and support both joint-space and end-effector-space control.
Robot-specific action dimensions are stored in embodiment configurations rather than hard-coded in policy adapters, which preserves a stable external schema across heterogeneous kinematic structures and control spaces.

Data-processing responsibilities are divided explicitly, and this division is load bearing.
The serving layer performs transport-level operations, including deserialization and image decoding, before observations reach the adapter.
The adapter performs only model-specific transformations, such as resizing, normalization, tokenization, temporal-context construction, and conversion from native predictions to standardized actions.
Fixing image decoding on the serving side means every adapter receives images in the same color order and layout, which removes a class of silent faults in which a policy is trained under one convention and evaluated under another.

\subsection{Integration and Conformance}
\label{sec:integration}

XPolicyLab treats policy integration as conformance to a shared specification rather than as environment-specific implementation work.
A policy package provides a runtime configuration, model adapter, execution scripts, documentation, and a reproducible checkpoint specification, with data-conversion and training entry points included when available.

Adapters are validated progressively, cheapest gate first.
Static checks verify configuration, imports, and script integrity.
An offline closed-loop client then exercises server startup, observation serialization, action structure, batched execution, and episode reset to a deterministic finish marker, without requiring a simulator or physical robot.
Only after these interface-level checks pass is the adapter connected to a supported environment.

Conversely, a new environment only needs to construct standardized observations, invoke the policy operations, and execute the returned actions, with policy-specific inference logic remaining behind the serving boundary.
Extending the ecosystem therefore requires one adapter per policy or one client per environment, rather than separate integration code for every pairing of policy and environment.

\subsection{Agent-Assisted Integration}
\label{sec:agent-skills}

Because the conformance procedure is fully specified, it can be executed rather than read.
The workflow is packaged as machine-readable \emph{agent skills} shipped with the repository: an integration skill that encodes the adapter layout, contract rules, checkpoint conventions, and the debug closed-loop procedure, and an audit skill that checks a submission against the same gates before review.
Coding agents such as Cursor, Claude Code, and Codex load these skills to scaffold, implement, and validate a new adapter under human supervision, turning policy integration from a documentation-reading exercise into a guided and checkable procedure.
Section~\ref{sec:integration-effort} quantifies the effect.

\section{Policy Ecosystem}
\label{sec:policy_ecosystem}

A central objective of XPolicyLab is to support heterogeneous robot policies without constraining their architectures, training objectives, or inference procedures.
As of August 8, 2026, XPolicyLab integrates \textbf{42 robot policies}, summarized in Table~\ref{tab:policy_ecosystem}: VLA models, world-action models, diffusion-based visuomotor policies, memory-augmented policies, and conventional imitation-learning baselines.

Each integration encapsulates model-specific dependencies, checkpoint loading, observation preprocessing, action decoding, and state management behind the common contract, so that environment clients invoke every supported policy through the same observation-update, action-prediction, batched-execution, and episode-reset operations.
All adapters expose the standardized serving and evaluation interface, while data-conversion and training entry points are included when the corresponding implementations are publicly available.

The ecosystem also spans heterogeneous temporal and computational requirements, from single-step prediction and action chunking to temporal aggregation and explicit memory, and from lightweight imitation-learning baselines to large foundation models that require an isolated GPU runtime.
Whether this breadth is genuinely absorbed by the contract, rather than merely tolerated, is an empirical question; Section~\ref{sec:integration-effort} answers it by measuring the effort that integration actually costs.

\section{Applications and Leaderboard Evidence}
\label{sec:apps}

\begin{figure}[h]
    \centering
    \includegraphics[width=0.95\linewidth]{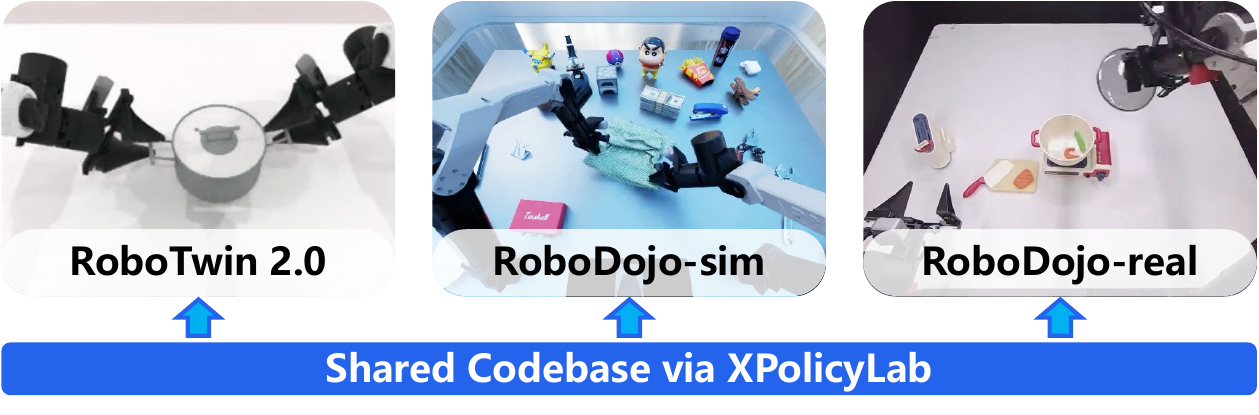}
    \caption{
\textbf{Cross-platform policy evaluation through XPolicyLab.}
A shared policy codebase and a standardized serving interface allow the same policy integration to be evaluated in RoboTwin 2.0, RoboDojo simulation, and RoboDojo real-world settings with minimal policy-side adaptation.
}
    \label{fig:shared_benchmark}
    \vspace{-10pt}
\end{figure}

We examine XPolicyLab through its deployment in two public benchmark ecosystems covering three evaluation settings (Fig.~\ref{fig:shared_benchmark}): RoboTwin simulation, RoboDojo simulation, and RoboDojo real-world evaluation.
These settings differ in task definitions, simulator infrastructure, robot embodiments, and evaluation protocols, yet share the same policy abstraction and serving architecture.
The leaderboards therefore characterize the operational coverage and portability of XPolicyLab rather than algorithmic improvements, and detailed task definitions and policy-level analyses belong to the corresponding benchmark papers~\cite{chen2025robotwin,chen2026robodojo}.

\subsection{RoboTwin Simulation}

RoboTwin evaluates bimanual manipulation across 50 tasks under clean and randomized settings~\cite{chen2025robotwin}.
XPolicyLab connects policies to RoboTwin through an environment client that constructs standardized observations and executes returned actions, while policy adapters remain responsible for model-specific preprocessing, checkpoint loading, internal state management, and action decoding.
This separation allows RoboTwin to reuse one serving stack across policies with heterogeneous architectures and dependencies.

Table~\ref{tab:robotwin_leaderboard} reports the top 10 policies ranked by mean success rate under the clean setting, with the randomized column characterizing performance under the more challenging configuration.
All listed policies completed evaluation on all 50 tasks under both settings.
The live leaderboard is available at \href{https://robotwin-platform.github.io/leaderboard}{robotwin-platform.github.io/leaderboard}.

\begin{table}[h]
\centering
\caption{Top 10 RoboTwin policies ranked by mean clean success rate. Snapshot dated Aug 10, 2026.}
\label{tab:robotwin_leaderboard}
\scriptsize
\setlength{\tabcolsep}{8pt}
\renewcommand{\arraystretch}{1.4}
\begin{tabular}{@{}clcc@{}}
\toprule
\textbf{Rank} &
\textbf{Policy} &
\textbf{Clean SR (\%)} &
\textbf{Randomized SR (\%)} \\
\midrule
1  & \XPLPolicyPNG{https://github.com/XPolicyLab/XPolicyLab/tree/main/policy/FastWAM}{Fast-WAM.png}{FastWAM}{yuan2026fast}                        & 77.8 & 1.9  \\
2  & \XPLPolicyPNG{https://github.com/XPolicyLab/XPolicyLab/tree/main/policy/Spatial_Forcing}{Spatial_Forcing.png}{Spatial Forcing}{li2026spatial} & 77.2 & 9.5  \\
3  & \XPLPolicyPNG{https://github.com/XPolicyLab/XPolicyLab/tree/main/policy/Pi_05}{Pi-05.png}{$\pi_{0.5}$}{intelligence2025pi_}                    & 70.7 & 46.0 \\
4  & \XPLPolicyPNG{https://github.com/XPolicyLab/XPolicyLab/tree/main/policy/X_WAM}{X-WAM.png}{X-WAM}{guo2026unified}                              & 70.0 & 25.8 \\
5  & \XPLPolicyPNG{https://github.com/XPolicyLab/XPolicyLab/tree/main/policy/X_VLA}{X-VLA.png}{X-VLA}{zheng2026x}                                  & 68.0 & 20.9 \\
6  & \XPLPolicyPNG{https://github.com/XPolicyLab/XPolicyLab/tree/main/policy/EventVLA}{EventVLA.png}{EventVLA}{yang2026eventvla}                   & 65.6 & 15.7 \\
7  & \XPLPolicyPNG{https://github.com/XPolicyLab/XPolicyLab/tree/main/policy/AHA_WAM}{AHA-WAM.png}{AHA-WAM}{cai2026aha}                            & 64.3 & 3.2  \\
8  & \XPLPolicyPNG{https://github.com/XPolicyLab/XPolicyLab/tree/main/policy/Xiaomi_Robotics_0}{Xiaomi-Robotics-0.png}{Xiaomi-Robotics-0}{cai2026xiaomi} & 62.9 & 18.2 \\
9  & \XPLPolicyPNG{https://github.com/XPolicyLab/XPolicyLab/tree/main/policy/GalaxeaVLA}{GalaxeaVLA.png}{G0}{jiang2025galaxea}                & 62.7 & 9.1  \\
10 & \XPLPolicyPNG{https://github.com/XPolicyLab/XPolicyLab/tree/main/policy/Abot_M0}{ABot-M0.png}{Abot-M0}{yang2026abot}                          & 57.4 & 22.9 \\
\bottomrule
\end{tabular}
\end{table}

\subsection{RoboDojo Simulation}

RoboDojo provides 42 simulation tasks organized into five capability dimensions: Generalization, Precision, Long-Horizon, Memory, and Open~\cite{chen2026robodojo}.
Its environment workers execute heterogeneous tasks concurrently in Isaac Sim, while XPolicyLab policy servers perform observation updates and batched action prediction.
This process separation matters most here, since many large policies carry dependencies that are incompatible with the simulator runtime.

Table~\ref{tab:robodojo_sim_leaderboard} reports the top 10 policies ranked by overall score averaged across the five capability dimensions.
Success rate (SR) measures complete task execution, whereas the score additionally credits partial task progress.
All evaluations use the common XPolicyLab environment-facing interface and the RoboDojo evaluation protocol.
The live leaderboard is available at \href{https://robodojo-benchmark.com/leaderboard}{robodojo-benchmark.com/leaderboard}.

\begin{table}[t]
\centering
\caption{Top 10 policies on the RoboDojo simulation leaderboard ranked by average score. Snapshot dated Aug 4, 2026.}
\label{tab:robodojo_sim_leaderboard}
\scriptsize
\setlength{\tabcolsep}{8pt}
\renewcommand{\arraystretch}{1.4}
\begin{tabular}{@{}clcc@{}}
\toprule
\textbf{Rank} &
\textbf{Policy} &
\textbf{Score} &
\textbf{SR (\%)} \\
\midrule
1  & \XPLPolicyPNG{https://github.com/XPolicyLab/XPolicyLab/tree/main/policy/G05}{GalaxeaVLA.png}{G0.5}{galaxea2026g05}                     & 20.23 & 14.88 \\
2  & \XPLPolicyPNG{https://github.com/XPolicyLab/XPolicyLab/tree/main/policy/Xiaomi_Robotics_1}{Xiaomi-Robotics-1.png}{Xiaomi-Robotics-1}{team2026xiaomi} & 20.07 & 13.93 \\
3  & \XPLPolicyPNG{https://github.com/XPolicyLab/XPolicyLab/tree/main/policy/Hy_Embodied_05_VLA}{Hy-Embodied-0.5-VLA.png}{Hy-Embodied-0.5-VLA}{zhang2026hy} & 13.07 & 8.80 \\
4  & \XPLPolicyPNG{https://github.com/XPolicyLab/XPolicyLab/tree/main/policy/Spatial_Forcing}{Spatial_Forcing.png}{Spatial Forcing}{li2026spatial} & 12.38 & 8.04 \\
5  & \XPLPolicyPNG{https://github.com/XPolicyLab/XPolicyLab/tree/main/policy/Pi_05}{Pi-05.png}{$\pi_{0.5}$}{intelligence2025pi_}                    & 11.41 & 6.91 \\
6  & \XPLPolicyPNG{https://github.com/XPolicyLab/XPolicyLab/tree/main/policy/InternVLA_A1_5}{InternVLA-A1.5.png}{InternVLA-A1.5}{ma2026internvla}  & 11.15 & 7.14 \\
7  & \XPLPolicyPNG{https://github.com/XPolicyLab/XPolicyLab/tree/main/policy/VLAct}{Spatial_Forcing.png}{VLAct}{ye2026starvla}                                                               & 10.66 & 7.60 \\
8  & \XPLPolicyPNG{https://github.com/XPolicyLab/XPolicyLab/tree/main/policy/X_VLA}{X-VLA.png}{X-VLA}{zheng2026x}                                  & 10.13 & 6.52 \\
9  & \XPLPolicyPNG{https://github.com/XPolicyLab/XPolicyLab/tree/main/policy/X_WAM}{X-WAM.png}{X-WAM}{guo2026unified}                              & 7.69  & 3.83 \\
10 & \XPLPolicyPNG{https://github.com/XPolicyLab/XPolicyLab/tree/main/policy/Xiaomi_Robotics_0}{Xiaomi-Robotics-0.png}{Xiaomi-Robotics-0}{cai2026xiaomi} & 6.93 & 4.18 \\
\bottomrule
\end{tabular}
\end{table}

\subsection{RoboDojo Real-World Evaluation}

RoboDojo-RealEval extends the same serving architecture to 18 physical tasks across three bimanual robot embodiments: ARX X5, Piper, and Piper X~\cite{chen2026robodojo}.
The robot controller acts as the XPolicyLab environment client, while policy inference runs in an isolated process, potentially on a remote GPU server.
Hardware control, sensing synchronization, scene reset, and safety supervision remain on the environment side.
When the required observation and action representations are preserved, a policy adapter is reused without modifying its model-specific inference implementation.

Each policy is evaluated over 10 trials per task, yielding 180 physical trials across the three embodiments.
Table~\ref{tab:robodojo_real_leaderboard} reports the top 10 policies ranked by overall score, excluding human teleoperation.

\begin{table}[t]
\centering
\caption{Top 10 policies on the RoboDojo-RealEval leaderboard ranked by average score. Snapshot dated Aug 4, 2026.}
\label{tab:robodojo_real_leaderboard}
\scriptsize
\setlength{\tabcolsep}{8pt}
\renewcommand{\arraystretch}{1.4}
\begin{tabular}{@{}clcc@{}}
\toprule
\textbf{Rank} &
\textbf{Policy} &
\textbf{Score} &
\textbf{SR (\%)} \\
\midrule
1  & \XPLPolicyPNG{https://github.com/XPolicyLab/XPolicyLab/tree/main/policy/Pi_05}{Pi-05.png}{$\pi_{0.5}$}{intelligence2025pi_}                    & 22.9 & 12.8 \\
2  & \XPLPolicyPNG{https://github.com/XPolicyLab/XPolicyLab/tree/main/policy/InternVLA_A1}{InternVLA-A1.png}{InternVLA-A1}{cai2026internvla}       & 12.0 & 7.2  \\
3  & \XPLPolicyPNG{https://github.com/XPolicyLab/XPolicyLab/tree/main/policy/GalaxeaVLA}{GalaxeaVLA.png}{G0}{jiang2025galaxea}                & 9.0  & 4.4  \\
4  & \XPLPolicyPNG{https://github.com/XPolicyLab/XPolicyLab/tree/main/policy/Xiaomi_Robotics_0}{Xiaomi-Robotics-0.png}{Xiaomi-Robotics-0}{cai2026xiaomi} & 7.9 & 3.9 \\
5  & \XPLPolicyPNG{https://github.com/XPolicyLab/XPolicyLab/tree/main/policy/X_VLA}{X-VLA.png}{X-VLA}{zheng2026x}                                  & 7.6  & 3.3  \\
6  & \XPLPolicyPNG{https://github.com/XPolicyLab/XPolicyLab/tree/main/policy/GR00T_N17}{GROOT-N1.7.png}{GR00T-N1.7}{bjorck2025gr00t}               & 5.9  & 1.7  \\
7  & \XPLPolicyPNG{https://github.com/XPolicyLab/XPolicyLab/tree/main/policy/Pi_0}{Pi-0.png}{$\pi_0$}{black2024pi_0}                               & 5.8  & 1.7  \\
8  & \XPLPolicyPNG{https://github.com/XPolicyLab/XPolicyLab/tree/main/policy/starVLA}{StarVLA.png}{StarVLA-$\alpha$}{community2026starvla}         & 4.1  & 1.7  \\
9  & \XPLPolicyPNG{https://github.com/XPolicyLab/XPolicyLab/tree/main/policy/Spirit_v15}{Spirit_v1.5.png}{Spirit v1.5}{spiritai2026spiritv15}      & 1.6  & 0.6  \\
10 & \XPLPolicyPNG{https://github.com/XPolicyLab/XPolicyLab/tree/main/policy/Dexbotic_DM0}{DM0.png}{Dexbotic-DM0}{yu2026dm0}                       & 0.0  & 0.0  \\
\bottomrule
\end{tabular}
\end{table}

Two observations matter for an infrastructure audience.
First, absolute performance is low in both simulation and reality, and real-world rankings only partially track simulation ones, so what currently binds progress is the cost of running a physical trial rather than the cost of another simulation seed.
Second, the gap between clean and randomized RoboTwin success rates differs sharply across policies, and that gap is interpretable only because all entries share one observation contract; had each submission shipped its own preprocessing, differences of this magnitude would be indistinguishable from packing discrepancies.

\section{System Evaluation}
\label{sec:system-evaluation}

\subsection{Integration and Reproduction Effort}
\label{sec:integration-effort}

We quantify integration cost with a controlled study on a representative task:
connecting $\pi_{0.5}$ to the RoboDojo simulation evaluation and reproducing a
closed-loop result.
Participants ($N=6$ engineers unfamiliar with XPolicyLab) performed the task
under three conditions: (i)~\emph{from scratch}, using only the upstream model
repository; (ii)~through XPolicyLab, following the adapter standard manually;
and (iii)~through XPolicyLab with the packaged agent skills, where a coding
agent executes the scaffold--implement--debug cycle under human supervision.

The agent condition fixes the agent stack so that the comparison is about the
skills rather than about model choice: all participants used Cursor as the
coding agent with Opus~5 as the underlying model, on the same repository
checkout, with the integration and audit skills loaded from
\texttt{.agents/skills/} and no additional prompting scaffold. The agent
performs scaffolding, adapter implementation, and iteration against the audit
gates, while the human supplies the run configuration, launches the debug
closed loop, and reviews the generated adapter before accepting it.

The study is within-subject: every participant completed all three conditions.
Six participants admit exactly the six orderings of three conditions, so each
ordering was assigned to one participant and no condition is systematically
advantaged by position. A participant's sessions were separated by at least one
day, and participants worked only from the materials of the current condition,
without reusing code or notes from an earlier one. Counterbalancing cannot
remove residual familiarity with $\pi_{0.5}$ itself, which is one reason we read
the result as an indicative case study rather than a controlled measurement of
the standard in isolation.
We measure time-to-first-successful-rollout and hand-written lines of code
(LoC), that is, code written by a human rather than generated by the agent and
reviewed, excluding checkpoint download time, which is bandwidth-bound and
identical across conditions.

\begin{table}[!tb]
\centering
\caption{Integration effort for connecting $\pi_{0.5}$ to RoboDojo simulation
(median over $N=6$ participants). The agent condition uses Cursor with Opus~5.}
\label{tab:integration-effort}
\footnotesize
\setlength{\tabcolsep}{0pt}
\renewcommand{\arraystretch}{1.15}
\begin{tabular*}{\columnwidth}{@{\extracolsep{\fill}}lccc@{}}
\toprule
\makecell[bl]{Phase}
  & \makecell[bc]{From\\scratch}
  & \makecell[bc]{XPolicyLab\\(manual)}
  & \makecell[bc]{XPolicyLab\\$+$ agent skills} \\
\midrule
Environment setup        & 2\,h      & 30\,min    & 10\,min \\
Obs./action glue code    & 2\,h      & 1\,h       & 10\,min \\
Debugging to 1st rollout & 1.5\,h    & 30\,min    & 10\,min \\
\midrule
\textbf{Total (median)}  & \textbf{$>$5\,h} & \textbf{$\sim$2\,h} & \textbf{$\sim$30\,min} \\
Hand-written LoC         & $\sim$300 & $\sim$120  & $\sim$0 \\
\bottomrule
\end{tabular*}
\end{table}

Table~\ref{tab:integration-effort} shows that conforming to the XPolicyLab
standard reduces the integration effort of a representative VLA policy from over
five hours to roughly two hours, and that the packaged agent skills reduce it
further to thirty minutes, an order-of-magnitude reduction over the from-scratch
baseline, because the agent executes the documented procedure while the human
only configures and reviews.
Reproducing the evaluation of an \emph{already-integrated} policy is cheaper
still: installation and a single evaluation command complete in $\sim$10
minutes, again excluding the bandwidth-bound checkpoint download, versus hours
of environment and glue-code reconstruction without the shared layer.

\section{Conclusion}

We presented \textbf{XPolicyLab}, a unified standard and open ecosystem for robot policy evaluation and deployment, combining a minimal policy adapter contract, standardized observation and action schemas, and dependency-isolated local and remote serving.
Measured across its 42 adapters, the design behaves as intended: model-specific code varies by an order of magnitude while the environment-facing loop stays within a few lines of a fixed reference, and conforming to the standard reduces the integration effort of a representative policy from over five hours to two hours, and to thirty minutes once the packaged agent skills execute the conformance procedure.
The same adapters drive RoboTwin, RoboDojo simulation, and RoboDojo-RealEval, so that heterogeneous policies retain their native implementations while sharing one evaluation pathway.
By replacing pairwise integration between policies and environments with reusable adapters and environment clients, XPolicyLab lowers the cost of the step that currently gates reproducible comparison: getting a checkpoint to run correctly in a setting where it has never run before.
The standard cannot remove physical variance or the scarcity of standardized robot time; it can only ensure that scarce time is spent on physical interaction rather than on rediscovering how to feed each checkpoint.

\begin{strip}
\section{Contributors}
\label{sec:contributor_list}

\footnotesize
\noindent
XPolicyLab is a collaborative open-source project led by
\textbf{MMLab@HKU} and \textbf{THU}, with project leadership provided by Tianxing Chen.
The complete list of contributors and their policy integrations is provided below.
Project website:
\href{https://xpolicylab.github.io/}{\texttt{xpolicylab.github.io}}.

\medskip
\noindent
\textbf{Core Lead Authors.}\\
Tianxing Chen, Yue Chen, Tian Nian, Zijian Cai, Guangyu Chen,
Wenwei Lin, Qiwei Liang and Zanxin Chen.

\medskip
\noindent
\textbf{Contributors.}\\
Peicheng Xiang, Kailun Su, Zixuan Li, Junyuan Tang, Yan Qin,
Qiangyu Chen, Shaolong Zhu, Xiang Li, Jiahao Zhang, Weijie Wan,
Baijun Chen, Honghao Su, Kehe Ye, Shujia Liu and Xspark AI Team
(\textbf{Benchmark Infrastructure};
\textbf{$\pi_{0.5}$},
\textbf{X-VLA},
\textbf{Xiaomi-Robotics-0},
\textbf{StarVLA},
\textbf{G0},
\textbf{ABot-M0},
\textbf{FastWAM},
\textbf{$\pi_0$},
\textbf{GR00T-N1.7},
\textbf{InternVLA-A1},
\textbf{SmolVLA},
\textbf{LDA-1B},
\textbf{MolmoAct2},
\textbf{GO-1},
\textbf{ACT},
\textbf{RDT-1B},
\textbf{DM0},
\textbf{TinyVLA},
\textbf{OpenVLA-OFT},
\textbf{RLDX-1});
Kaixuan Wang, Haotian Liang (\textbf{Hy-Embodied-0.5-VLA}, \textbf{RxBrain});
Yunze Liu (\textbf{DreamZero});
Mingleyang Li and Yuran Wang (\textbf{Mem-0});
Boyu Chen (\textbf{UniT});
Hongzhe Bi (\textbf{Motus H-RDT});
Shuhe Huang and Hengkai Tan (\textbf{MotuBrain});
Jisong Cai and Yao Mu (\textbf{AHA-WAM});
Jun Guo (\textbf{X-WAM});
Xiaofeng Wang, Zheng Zhu, Weijie Ke, and Hengtao Li
(\textbf{GigaWorld-Policy});
Yuhang Tang and Xiaofan Li (\textbf{Wall-WM}, \textbf{Wall-OSS});
Ganlin Yang and Zhangzheng Tu (\textbf{EventVLA});
Shuai Yang (\textbf{LingBot-VLA}, \textbf{LingBot-VA});
Yiqing Wang and Tengyue Jiang (\textbf{Being-H0.5} and \textbf{VITRA})
Wenxuan Song and Pengxiang Ding (\textbf{Spatial Forcing});
Kaidong Zhang and Yu Sun (\textbf{A1});
Junliang Guo, Tong Zhang, and Yixing Chen (\textbf{Spirit v1.5});
Rongxu Cui and Zongzheng Zhang (\textbf{Dexora-1B});
Haoxiang Ma and Junhao Cai (\textbf{InternVLA-A1.5});
Haoyu Zhang (\textbf{G0.5});
Senqiao Yang, Jinhui Ye, Pengguang Chen, and Shu Liu (\textbf{VLAct});
Xiu Su, Wenhan Fang, Wenhao Li, Yichao Cao, Chengyao Wang,
and Qiang Chen (\textbf{CSU-AI-0}); XR-1 Team (\textbf{Xiaomi-Robotics-1}).

\medskip
\noindent
\textbf{Corresponding Authors.}\\
Wenbo Ding and Ping Luo.
For questions regarding the project, please contact
\href{mailto:ding.wenbo@sz.tsinghua.edu.cn}
{\texttt{ding.wenbo@sz.tsinghua.edu.cn}} and \href{mailto:pluo@cs.hku.hk}{\texttt{pluo@cs.hku.hk}}.

\medskip
\noindent
XPolicyLab is being actively maintained and expanded; we will update this report as the project evolves.


\normalsize
\end{strip}





\bibliographystyle{IEEEtran}
\bibliography{main}

\end{document}